\documentclass[10pt,twocolumn,letterpaper]{article}

\usepackage{wacv}
\usepackage{times}
\usepackage{epsfig}
\usepackage{graphicx}
\usepackage{amsmath}
\usepackage{amssymb,bm}
\usepackage{subfigure}
\usepackage{multirow}
\usepackage{pifont}
\usepackage{dsfont}
\usepackage[mode=text]{siunitx} 

\newcommand{\xmark}{\ding{55}}
\newcommand{\name}{SGSG}
\newcommand{\obs}{\text{obs}}
\newcommand{\pred}{\text{pred}}
\newcommand{\gt}{{\text{gt}}}

\newcommand{\ol}{T_{\obs}}
\newcommand{\pl}{T_{\pred}}
\newcommand{\diag}{\text{diag}}

\def\wacvPaperID{256} 

\wacvfinalcopy 

\ifwacvfinal
\def\assignedStartPage{1} 
\fi

\ifwacvfinal
\usepackage[breaklinks=true,bookmarks=false]{hyperref}
\else
\usepackage[pagebackref=true,breaklinks=true,colorlinks,bookmarks=false]{hyperref}
\fi

\ifwacvfinal
\else
\fi

\begin{document}
\setlength\abovedisplayskip{1pt}
\setlength\belowdisplayskip{1pt}

\title{Scene Gated Social Graph: Pedestrian Trajectory Prediction \\ Based on Dynamic Social Graphs and Scene Constraints}

\author{Hao Xue \hspace{2cm} Du Q. Huynh \hspace{2cm} Mark Reynolds\\
The University of Western Australia\\
{\tt\small hao.xue@research.uwa.edu.au \hspace{2cm} \{du.huynh, mark.reynolds\}@uwa.edu.au}
}

\maketitle

\begin{abstract}
Pedestrian trajectory prediction is valuable for understanding human motion behaviors and it is challenging because of the social influence from other pedestrians, the scene constraints and the multimodal possibilities of predicted trajectories.
Most existing methods only focus on two of the above three key elements.
In order to jointly consider all these elements, we propose a novel trajectory prediction method
named \textbf{S}cene \textbf{G}ated \textbf{S}ocial \textbf{G}raph (\name).
In the proposed \name, dynamic graphs are used to describe the social relationship among pedestrians.
The social and scene influences are taken into account through the scene gated social graph features which combine the encoded social graph features and semantic scene features.
In addition, a VAE module is incorporated to learn the scene gated social feature and sample latent variables for generating multiple trajectories that are socially and environmentally acceptable.
We compare our \name\ against twenty state-of-the-art pedestrian trajectory prediction methods and the results show that the proposed method achieves superior performance on two widely used trajectory prediction benchmarks.
\end{abstract}

\section{Introduction}
\label{sec:intro}
The forecasting of pedestrian trajectories is essential for numerous applications ranging from autonomous robots for delivery and driverless vehicles for collision avoidance to crowd management from a security aspect. 
From the pioneering Social Force Model based approaches~\cite{helbing1995social,pellegrini2009you,yamaguchi2011you} 
to data-driven Long Short Term Memory (LSTM) based methods~\cite{Alahi_2016_CVPR,hao_2017_dicta,ACMMM17_YukeLi_pedestrian,fernando2018soft,Bhattacharyya_2018_CVPR,xue2019location,Gupta_2018_CVPR,Hasan_2018_CVPR,Xu_2018_CVPR,zhang2019sr,Chandra_2019_CVPR,Li_2019_CVPR,xue2019pedestrian,Ivanovic_2019_ICCV,Rasouli_2019_ICCV,xue2020poppl,Sun_2020_CVPR} or Convolutional Neural Network (CNN) based methods~\cite{yi2016pedestrian,nikhil2018convolutional,Deo_2018_CVPR_Workshops,Makansi_2019_CVPR}, many models have been designed for the prediction of pedestrians trajectories.
However, it is still a challenging task because of three key elements that need to be addressed in trajectory prediction:

\textbf{\textit{(i) Social Influence.}}
One of the most significant current discussions in pedestrian trajectory prediction is \textit{person-person interaction}, also known as \textit{social influence}. When people move in public areas, they exhibit different social behaviors: keeping the same pace when they walk in a group, turning slightly to avoid collision with incoming pedestrians, etc. Such person-person interaction plays a major role in the pedestrians' trajectories in a scene. 

\textbf{\textit{(ii) Multimodal Nature of Trajectory Prediction.}}
There are several plausible future paths for a pedestrian to move around in a scene. 
For example, a person can choose to turn right or turn left to bypass an obstacle.

\textbf{\textit{(iii) Scene Constraints.}}
The physical environment surrounding pedestrians is another key element that should not be ignored. There are static obstacles and building structure that constrain the trajectories of pedestrians. So trajectory prediction methods that consider only social influence may end up generating trajectories that are not environmentally acceptable. 

The importance of these three key elements is not unrecognized in the literature. Indeed, many existing studies have given very thorough treatment to social influence in pedestrian trajectory prediction. The social LSTM model~\cite{Alahi_2016_CVPR} is one of the early papers that uses a neighborhood region around each person of interest (POI) to capture social influence information.
Other methods such as \cite{vemula2017social,sun2019socially} use complete graphs to model social influence.
While these papers and other related studies~\cite{AAAI18_SAGAIL,fernando2018soft,Hasan_2018_CVPR,zhang2019sr,xue2019pedestrian} analyze social influence in depth, they have not integrated scene constraints and multimodal predictions into their models.

For the scene constraint element, several existing studies such as~\cite{xue2018ss,syed2019sseg,lisotto2019social} have incorporated it as well as the social influence into their work. For example, a hierarchical LSTM network called SS-LSTM~\cite{xue2018ss} that integrates scene context extracted by a CNN and social influence has been used in trajectory prediction.
In addition to the social pooling for capturing social influence, a semantic pooling based on pre-defined pixel-level semantic maps has been designed to capture scene influence~\cite{lisotto2019social}. However, these methods have not been designed to capture the multimodal nature of pedestrian trajectories.

In the trajectory prediction task, multimodality is commonly tackled by designing a suitable latent variable $z$ through which random samples are drawn, often from the normal distribution ${\cal N}(0,1)$. Representative work includes the Social GAN~\cite{Gupta_2018_CVPR} which extends the Social LSTM model~\cite{Alahi_2016_CVPR} by adding multimodal predictions and the STGAT method~\cite{huang2019stgat} which uses the ${\cal N}(0,1)$ sampling with their spatial-temporal graph attention network. Instead of the normal distribution, the mean $\bm{\mu}$ and standard deviation ${\bm\sigma}$ of an appropriate dimension have also been learned for the latent variable ${\bf z}$. Typical examples that use the ${\cal N}({\bm\mu},\diag({\bm\sigma}^2))$ distribution for multimodality include the IDL~\cite{Li_2019_CVPR} and Trajectron~\cite{Ivanovic_2019_ICCV}. While the above methods handle multimodal predictions, they have not included both social influence and scene constraints. 

There are only a few existing pedestrian trajectory predictors that consider all the three elements.
DESIRE~\cite{Lee_2017_CVPR} adopts a conditional variational auto-encoder (CVAE) and an RNN to learn $\bm{\mu}$ and $\bm{\sigma}$ for the latent variable. 
It has a scene context fusion module to capture scene features and social interactions of pedestrians.
SoPhie~\cite{Sadeghian_2019_CVPR}, on the other hand, uses a VGGNet-19~\cite{vgg_2015} and two attention modules to encode the scene and social influences. 
For multimodal predictions, it adopts a GAN based model. 
Based on BicycleGAN~\cite{zhu2017toward}, Social-BiGAT~\cite{socialbigat_neurips19} develops a bijection between the output trajectories and the latent space input to the trajectory generator.
It utilizes a graph attention network and VGG to encode social and scene influences.

To tackle all the three key elements described above, we propose a novel architecture which we call \name\ (\textit{\textbf{S}cene \textbf{G}ated \textbf{S}ocial \textbf{G}raph}).
Our \name\ differs, in two respects, from the aforementioned pedestrian trajectory predictors that also implement all the three elements.
\textbf{First}, although both DESIRE~\cite{Lee_2017_CVPR} and our method use VAE to generate multiple trajectories, they are different in the way scene and social influences are incorporated.
DESIRE~\cite{Lee_2017_CVPR} uses the encoded hidden states of the POI's own trajectory as input to the VAE, so $\bm{\mu}$ and $\bm{\sigma}$ are learned from the POI only, whereas \name\ directly passes social graph and scene features to the VAE, so the learned $\bm{\mu}$ and $\bm{\sigma}$ capture these influences.
As a result, DESIRE~\cite{Lee_2017_CVPR} requires an extra refinement stage to incorporate social and scene influences.
Our proposed way of learning ${\bm\mu}$ and ${\bm\sigma}$ also differs from Social-BiGAT~\cite{socialbigat_neurips19} in that their estimates of these parameters do not include scene context information. 
\textbf{Second}, for the social influence element, we propose to use dynamic star graphs to encode other pedestrians for each POI. In contrast to existing studies~\cite{vemula2017social,sun2019socially,huang2019stgat,Mohamed_2020_CVPR} which commonly use the complete graph structure, our star graph structure is memory efficient.
Our social encoder using dynamic star graphs also has an advantage over SoPhie~\cite{Sadeghian_2019_CVPR} in that it does not impose a limit to the maximum number of pedestrians in the scene.

In summary, the contributions of this paper are:
(i)~We propose a novel scene gated social feature to incorporate scene and social cues to trajectory prediction.
Through our ablation studies which evaluate the proposed gating operation against other feature merging methods, such as concatenation and addition, the superiority of our method is validated.
(ii)~We design a dynamic star graph data structure to model the social graph features around pedestrians. 
Our experimental evaluation demonstrates that these star graphs are not only memory efficient but also effective in producing state-of-the-art prediction results.
(iii)~Through exhaustive experiments, we show that~\name\ achieves state-of-the-art performances in predicting both single and multimodal future trajectories.

\section{Related Work}
\label{sec:2}
Many methods in the literature of pedestrian trajectory prediction focus on forecasting trajectories that can better cope with the effect of social influence~\cite{yamaguchi2011you,Alahi_2016_CVPR,Xu_2018_CVPR,zhang2019sr,Ivanovic_2019_ICCV} (person-person interactions), surrounding physical environment~\cite{bartoli2017context,hao_2017_dicta,manh2018scene,ridel2019scene,syed2019sseg} (person-scene interaction), or both types of interactions~\cite{Lee_2017_CVPR,xue2018ss,Liang_2019_CVPR,Sadeghian_2019_CVPR}. With the generative models such as the Generative Adversarial Network (GAN)~\cite{Goodfellow-etal-NIPS14} demonstrating their successes in other application areas~\cite{yu2017seqgan,choi2018stargan,kupyn2018deblurgan}, a few methods~\cite{Gupta_2018_CVPR,Sadeghian_2019_CVPR,Amirian_2019_CVPR_Workshops,Li_2019_CVPR,socialbigat_neurips19,huang2019stgat} start to incorporate multimodality into trajectory prediction.

\noindent \textbf{Person-Person Interaction}
A classical paper that models person-person interaction is the Social Force Model (SFM)~\cite{helbing1995social} which uses the attractive and repulsive forces to capture the social influence among pedestrians.
Following the SFM, Pellegrini~\etal~\cite{pellegrini2009you} propose the Linear Trajectory Avoidance (LTA) algorithm to minimize collisions among pedestrians and to also improve pedestrian tracking; Yamaguchi~\etal~\cite{yamaguchi2011you}, on the other hand, include more behavior factors such as damping and collision to model social interactions.
In addition to the Social LSTM described in the previous section,
the Social Attention method~\cite{vemula2017social} and the Attention LSTM method~\cite{fernando2018soft} take advantages of the attention mechanism and imposed different degrees of attention to different neighbors in the trajectory prediction network. While the MX-LSTM method~\cite{Hasan_2018_CVPR} takes the head pose information into account during trajectory forecasting, SR-LSTM~\cite{zhang2019sr} uses a social-aware information selection mechanism for message passing between neighboring pedestrians.
Another recent method involves using a social pyramid~\cite{xue2019pedestrian} to differentiate the influences from pedestrians in nearby neighborhoods and those in remote areas. All the methods above do not model scene influence and can not generate multiple trajectory predictions.

\noindent \textbf{Person-Scene Interaction}
Pedestrians interact with the scene by avoiding static objects lying along their walking paths.
Bartoli~\etal~\cite{bartoli2017context} model the influence of static objects in the scene in their trajectory prediction method through a context-aware pooling strategy.
Syed and Morris~\cite{syed2019sseg} replace the scene context part of SS-LSTM by the semantic segmentation architecture from  SegNet~\cite{badrinarayanan2017segnet}.
Rather than dealing with the whole scene uniformly, the CAR-Net model of Sadeghian~\etal~\cite{Sadeghian_2018_ECCV} uses raw scene images and targets at specific areas when predicting trajectories.
RSBG~\cite{Sun_2020_CVPR} extracts scene features of a POI through a sequence of image patches that contain the POI.
To retrieve detailed spatial information of the scene, the Scene-LSTM method~\cite{manh2018scene} divides the scene into small grid-cells which are further divided into sub-grids.
Based on the relation reasoning network~\cite{santoro2017simple}, Choi and Dariush~\cite{Choi_2019_CVPR_Workshops,Choi_2019_ICCV} propose to use a gated relation encoder to discover both person-person interaction and person-scene interaction relations.
Liang~\etal~\cite{Liang_2019_CVPR} use a person behavior module and a person interaction module in their LSTM network to capture information on pedestrians' behaviors and their interaction with the surroundings. 
While most of the methods reviewed above also incorporate person-person interaction, they simply concatenate scene features and social features together. This is different from our \name\ method where dynamic social graphs are constructed for each POI and then combined with the scene features through a `gating' operation (details in Section~\ref{sec:scene}) for the downstream operation.

\noindent \textbf{Multimodality}
Based on the GAN architecture, Gupta~\etal~\cite{Gupta_2018_CVPR} propose the Social GAN model (abbreviated as SGAN) for generating multiple trajectory predictions. It uses a pooling module to expand the neighborhood around each POI to the whole scene so that all the pedestrians are covered.
Similar to the SGAN, Social Ways~\cite{Amirian_2019_CVPR_Workshops}, CGNS~\cite{Jiachen_IROS19}, MATF~\cite{Zhao_2019_CVPR}, IDL~\cite{Li_2019_CVPR}, PMP-NMMP~\cite{Hu_2020_CVPR}, and Sun~\etal~\cite{Sun1_2020_CVPR} also take advantage of the GAN to generate trajectories.
SoPhie, proposed by Sadeghian~\etal~\cite{Sadeghian_2019_CVPR}, is another GAN based trajectory prediction model.
As described in Section~\ref{sec:intro}, SoPhie incorporates social and scene influences and can handle multimodality.
Following SoPhie, Social-BiGAT~\cite{socialbigat_neurips19} makes use of the BicycleGAN~\cite{zhu2017toward} to produce multiple predictions for each observed trajectory.
Instead of using GAN, Zhang~\etal~\cite{zhang2019stochastic} propose to use a temporal stochastic method to sequentially learn a prior model of uncertainty during prediction.
The STGAT method~\cite{huang2019stgat} of Huang~\etal, on the other hand, produces multiple predicted trajectories by randomly sampling a latent variable $\mathbf{z}$ from $\mathcal{N}(0, 1)$.
IDL~\cite{Li_2019_CVPR} takes it further by learning $\bm{\mu}$ and $\bm{\sigma}$ from the motion features constructed from pedestrians' trajectories.

\section{Proposed Method}
\label{sec:3}
\subsection{Problem Formulation and System Overview}
In pedestrian trajectory prediction, we assume that the locations of each pedestrian are acquired through a person detection and tracking system in advance.
The 2D coordinate $(x_i^t, y_i^t) \in \mathds{R}^2$ stands for the location of $i^\text{th}$ POI at time $t$.
Depending on the data, coordinates can be in meters (world coordinates) or in pixels (image coordinates).
At the time step $\ol$, the history locations of each POI from $t=1$ to $t=\ol$ are observed. The goal of pedestrian trajectory prediction is to forecast the future locations of the POI from $t={\ol}\!+\!1$ to $t={\ol}\!+\!{\pl}$.
To avoid confusion, in the rest of this paper, $t=1,\cdots,\ol$ and $t={\ol}\!+\!1,\cdots, {\ol}\!+\!{\pl}$ are referred as the \textit{observation period} and the \textit{prediction period}.
We denote the $i^\text{th}$ observed trajectory by  ${\bf{X}}_{\obs}^{i}\!=\! \big[\! \left(x_{1}^{i},y_{1}^{i}\right), \cdots, \left(x_{\ol}^{i},y_{\ol}^{i}\right)\!\big]$, the ground truth trajectory by ${\bf{X}}_{\gt}^{i} = \big[ \left (x_{{\ol}+1}^{i},y_{{\ol}+1}^{i} \right), \cdots ,( x_{{\ol} + {\pl}}^{i},y_{{\ol} + {\pl}}^{i} )\big]$, and the predicted trajectory by
${\hat{\bf{X}}}_{\pred}^{i} = \big[(\hat{x}_{{\ol}+1}^{i},\hat{y}_{{\ol}+1}^{i}), $ $\cdots ,( \hat{x}_{{\ol} + {\pl}}^{i},\hat{y}_{{\ol} + {\pl}}^{i})\big]$.

\begin{figure*}[!tbp]
  \centering
  \includegraphics[width=.843\textwidth]{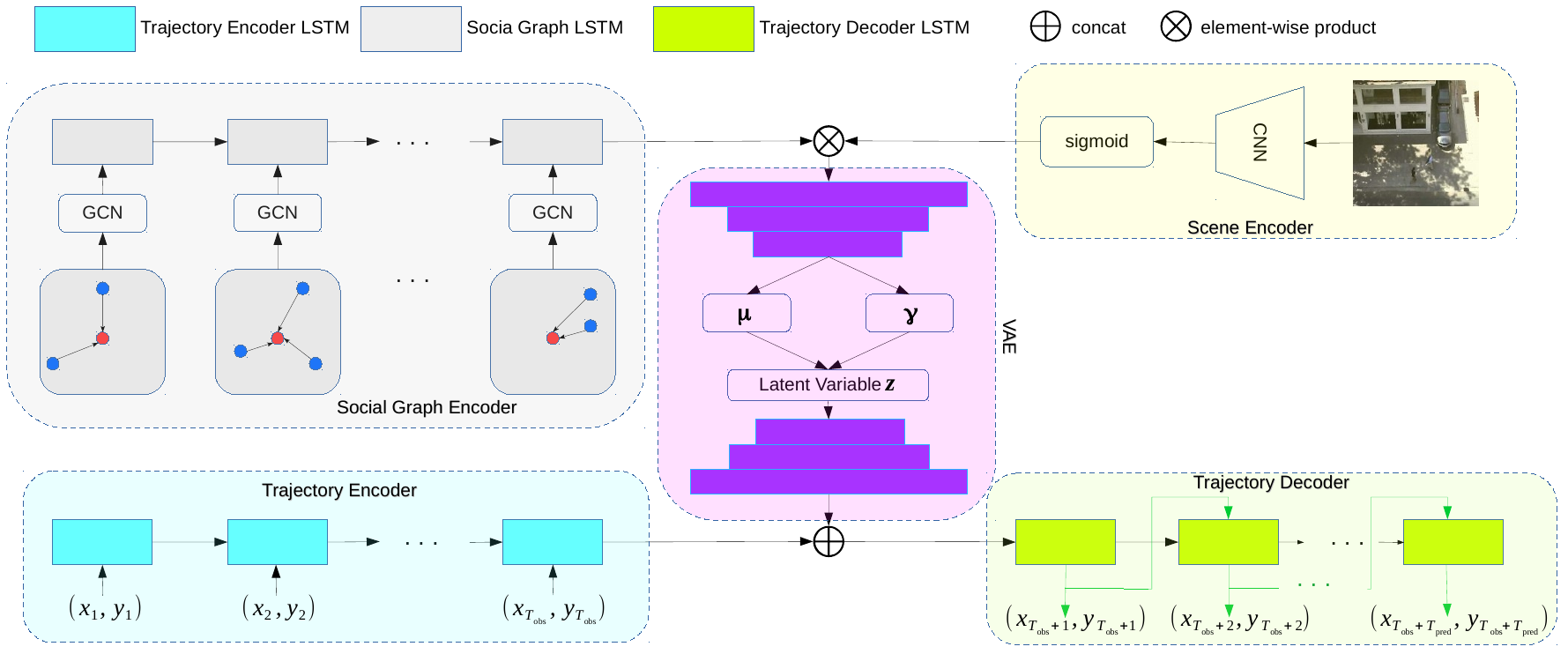} \\ 
  \caption{The architecture of \name. Two LSTM based encoders are used to encode social graphs and the POI's own history trajectory separately. 
  The scene image is processed through the scene encoder module.
  \name\ also includes a VAE to sample the latent variable ${\bf z}$ which is used to reconstruct the scene gated social graph feature for multimodal trajectory prediction through the trajectory decoder.
  To simplify the visualization, the embedding vectors ${\bf e}_t$ for all $t$, are not shown.
  }
\label{fig:SGS}
\end{figure*}

Our \name\ model~(Fig.~\ref{fig:SGS}) consists of:
(i)~a social graph encoder designed to extract social graph features (Section~\ref{sec:sg});
(ii)~a scene encoder module to extract scene features to tackle the scene constraints (Section~\ref{sec:scene});
(iii)~a VAE module to learn the latent variable $\mathbf{z}$ and reconstruct the scene gated social graph features (Section~\ref{sec:prediction});
(iv)~a trajectory encoder-decoder architecture for encoding each POI's own history trajectory and generating future trajectories (Section~\ref{sec:prediction}).

\subsection{Social Graph Encoder}\label{sec:sg}

\begin{figure}[!tbp]
  \centering
  \subfigure{
  \includegraphics[width=1.3cm]{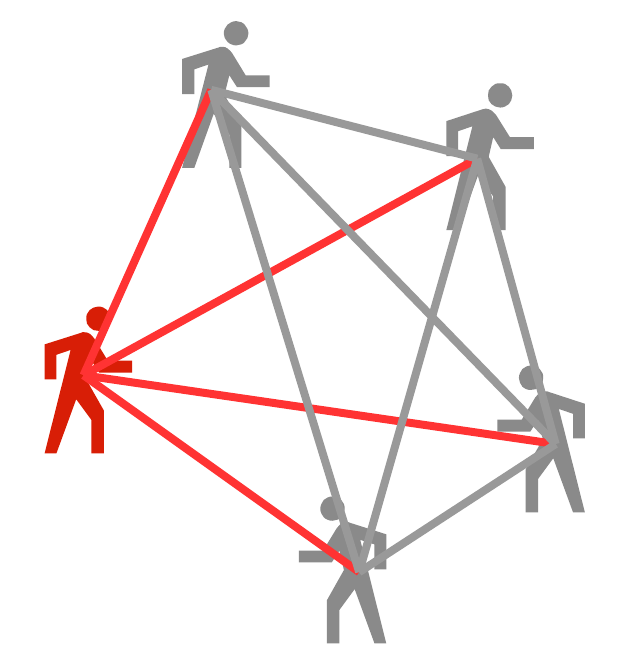}} \hspace{4ex}
  \subfigure{
  \includegraphics[width=1.3cm]{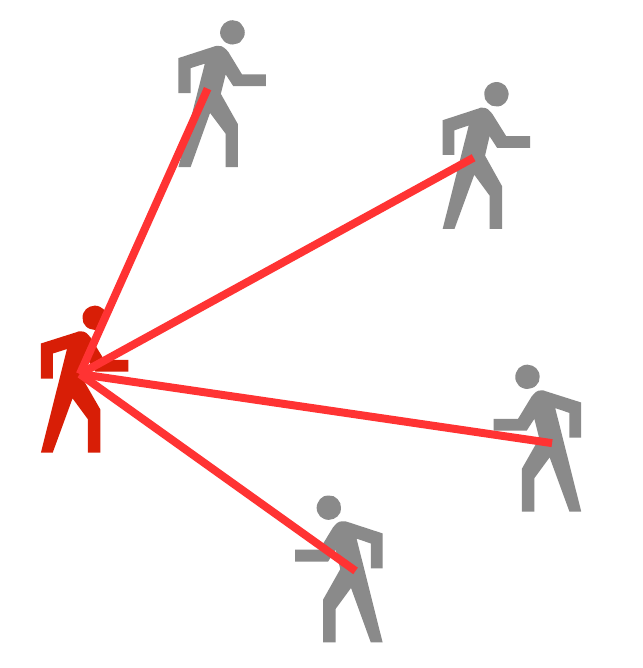}} \\ 
  \caption{Two types of pedestrian graphs: complete graph (left) and star graph (right). The star graph is used in the proposed social graph encoder as it focuses on the influence from other pedestrians (gray) to the POI (red).}
  \label{fig:graphs}
\end{figure}

The Graph Neural Network was introduced in~\cite{scarselli2009graph} and can be defined as an ordered pair $\mathcal{G}=(V,E)$, where
$V$ and $E$ are the sets of nodes and edges that link the nodes.
For the trajectory prediction task, it is natural to consider the pedestrians in a scene as nodes of the graph at each time step.
In this paper we model pedestrians as the only dynamic entities in the scene. For more complicated scenarios where the location coordinates of other moving objects, such as vehicles, are also available, they can be included in the  model.
More specifically, we use a dynamic graph $\mathcal{G}_i^t=(V_i^t, E_i^t)$ to describe the social influence of the $i^{\text{th}}$ POI at time $t$.
The set of nodes $V_i^t$ consists of the POI and a set of all other pedestrians, denoted by $\mathcal{N}_{(i)}^t$, in the scene at the same time $t$.
An edge is then assigned between the $i^{\text{th}}$ POI and each neighbor $j$, for $j\in \mathcal{N}_{(i)}^t$, to form $E_i^t$.
As the number of neighbours for each POI at each time instant is variable, the social graph does not have a fixed architecture.
This constructed star graph is different from the complete graph used in STGAT~\cite{huang2019stgat} (see Fig.~\ref{fig:graphs}). With fewer edges, our star graph is shown to be more efficient computationally yet equally effective for modeling the social interaction of the POI with other pedestrians.

A graph convolutional network (GCN in Fig.~\ref{fig:SGS}) is applied to encode each constructed social graph $\mathcal{G}_i^t$.
Mathematically, the graph convolutional operation can be  described as follows. The initial embedding social graph feature $\mathbf{a}_{i,0}^{t}$ at layer $0$ can be set to the location coordinates of the POI. \ie,
$
    \mathbf{a}_{i,0}^{t} = (x_{i}^{t}, y_{i}^{t}).
$
Given $\mathbf{a}_{i,l}^{t}$ at layer $l$, the embedding feature at layer $l+1$ is defined as
\begin{equation}
    \mathbf{a}_{i,l+1}^{t} = \text{ReLU}\biggl(\mathbf{b}^l + \frac{1}{ | \mathcal{N}_{(i)}^t |}\sum_{j\in \mathcal{N}^t_{(i)}}\mathbf{W}^l \mathbf{a}_{j,l}^{t}\biggr), \label{eq:social_graph}
\end{equation}
where $\mathbf{W}^l$ and $\mathbf{b}^l$ are the trainable weight matrix and bias term in the $l^{\text{th}}$ graph convolutional layer.

To further capture the temporal social information during the observation period, our \name\ network uses an LSTM layer, denoted by $\text{LSTM}_{\text{SG}} (\cdot)$ in Eq.~\eqref{eq:social_lstm} below, to encode the embedding social graph features $\mathbf{a}_{i}^{t}$ (the subscript $l$ has been dropped to simplify the explanation) across the time steps of the observation period:
\begin{equation}
    \mathbf{g}_{i}^{t} = \text{LSTM}_{\text{SG}}(\mathbf{g}_{i}^{t-1}, \mathbf{a}_{i}^{t};\mathbf{W}_{\text{SG}}). \label{eq:social_lstm}
\end{equation}
Here, $\mathbf{W}_{\text{SG}}$ represents the trainable parameters in this LSTM layer.

\subsection{Scene Encoder}\label{sec:scene}

The scene encoder of \name\ takes a scene image $I_i^{\ol}$ as input and passes it through a CNN to yield the visual feature $\mathbf{s}_{i}^{\ol}$:
\begin{equation}
    \mathbf{s}_i^{\ol} = \text{CNN}(I_i^{\ol}; \mathbf{W}_{\text{CNN}}).
    \label{eq:cnn}
\end{equation}
Unlike SoPhie~\cite{Sadeghian_2019_CVPR} and Social-BiGAT~\cite{socialbigat_neurips19} that used a VGG~\cite{vgg_2015} as the CNN feature extractor to extract scene features, we use the encoder module of the semantic segmentation architecture DeepLabv3+~\cite{Chen_2018_ECCV} to extract semantic features.
The encoded semantic features are then passed through a convolutional layer and a single fully connected layer to form the scene features $\mathbf{s}_i^{\ol}$, $\forall i$.
The encoder module of DeepLabv3+ used in our scene encoder is initialized with the weight matrix that has been pre-trained on Cityscapes~\cite{Cordts2016Cityscapes}.
Our scene encoder (the yellow box in Fig.~\ref{fig:SGS}) is then trained together with other parts of~\name.

For each POI, only the frame at $t=\ol$ is used to extract the semantic scene feature $\mathbf{s}_i^{\ol}$. This is due to the following two observations:
(i)~The scene at $t=\ol$ is more important and contains more up-to-date scene information than the history frames where $t < \ol$; ~(ii)~Since pedestrians (the only dynamic objects currently modeled by \name) are already handled by the Social Graph Encoder and since static objects in the scene, like trash bins, water fountains, etc, have fixed location coordinates for all $t$, it is not necessary to pass all video frames to the scene encoder. 
In terms of training \name\ to learn the scene constraints, as different pedestrians' trajectories cover different periods of timestamps in the video, there are sufficient scene images passed to the scene encoder for training (around 63\% of annotated video frames in the experiments are used in the training phase).

The social graph features and scene features capture the dynamic and static aspects of the scene. To combine these features for the downstream operation of~\name, we propose to use a scene gated social graph feature $\mathbf{G}_{i}^{\ol}$ defined as:
\begin{equation}
    \mathbf{G}_{i}^{\ol} =   \text{sigmoid} (\mathbf{s}_{i}^{\ol}) \otimes \mathbf{g}_{i}^{\ol}, \label{eq:social_scene}
\end{equation}
where the $\otimes$ represents element-wise product. The sigmoid function applied to the scene feature $\mathbf{s}_{i}^{\ol}$ non-linearly scales it to the range (0,1) and has the effect as a gate on the encoded temporal social graph features $\mathbf{g}_{i}^{\ol}$. 
Motivated by the gating mechanism inside the LSTM architecture, this gating process is designed to use the scene features to control the passing or rejection of social graph features.

\subsection{Multimodal Predictions}\label{sec:prediction}
Before making predictions about the future routes, we need the knowledge about the past movement information of the POI.
To this end, a trajectory encoder $\text{LSTM}_{\text{ENC}}(\cdot)$ is used to encode the POI's own history trajectory in the observation period through:
\begin{align}
    \label{eq:emb}
    \mathbf{e}_{i}^{t} &= \phi  (x_{i}^{t}, y_{i}^{t}; \mathbf{W}_{\text{EMB}}),\\
    \label{eq:lstm_enc}
    \mathbf{h}_{i}^{t} &= \text{LSTM}_{\text{ENC}}(\mathbf{h}_{i}^{t-1}, \mathbf{e}_{i}^{t};\mathbf{W}_{\text{ENC}}),
\end{align}
where $\phi (\cdot)$ is an embedding layer, $\text{LSTM}_{\text{ENC}}$ is the LSTM layer in the trajectory encoder, and
$\mathbf{W}_{\text{EMB}}$ and $\mathbf{W}_{\text{ENC}}$ are the associated weight matrices.

To achieve multimodal predictions, we use a VAE as the generative network in the proposed \name\ to sample the scene gated social graph features via:
\begin{align}
    \bm{\mu}_i, \bm{\gamma}_i &= \text{VAE}_{\text{enc}} (\mathbf{G}_{i}^{\ol}; \mathbf{W}_{\text{venc}}),\\
    \label{eq:sample}
    \mathbf{z}_i &\sim Q(\mathbf{z}_i | \mathbf{G}_{i}^{\ol}) = \mathcal{N}(\bm{\mu}_i, \diag(\bm{\sigma}_i^2)),\\
    \label{eq:vae_decoder}
    \mathbf{\hat{g}}_{i}^{\ol} &=  \text{VAE}_{\text{dec}} (\mathbf{z}_i; \mathbf{W}_{\text{vdec}}),
\end{align}
where $\text{VAE}_{\text{enc}}(\cdot)$, the encoding part of the VAE, outputs the mean vector $\bm{\mu}_i$ and the logarithmic variance $\bm{\gamma}_i$ (\ie, $\bm{\gamma}_i \triangleq \log\bm{\sigma}_i^2$) of a Gaussian distribution, from which a latent variable $\mathbf{z}_i$ is sampled.
Based on $\mathbf{z}_i$, a reconstructed scene gated social graph feature $\mathbf{\hat{g}}_{i}^{\ol}$ is then calculated through $\text{VAE}_{\text{dec}} (\cdot)$, the decoding part of the VAE.

In the prediction phase (${\ol}\!+\!1 \leqslant t \leqslant {\ol}\!+\!{\pl}$), we initialize the hidden state $\hat{\mathbf{h}}_{i}^{\ol}$ of $\text{LSTM}_{\text{DEC}}$ as:
\begin{equation}
    \hat{\mathbf{h}}_{i}^{\ol} = \mathbf{\hat{g}}_{i}^{\ol} \oplus \mathbf{h}_{i}^{\ol}, \label{eq:dec_init}
\end{equation}
where $\oplus$ denotes the concatenation operation.
After that, a predicted trajectory is generated by computing Eq.~\eqref{eq:dec_emb} to Eq.-\eqref{eq:out} recurrently:
\begin{align}
\label{eq:dec_emb}
\hat{\mathbf{e}}_{i}^{t} &= \phi  (\hat{x}_{i}^{t}, \hat{y}_{i}^{t}; \mathbf{W}_{\text{EMB}}), \\
    \hat{\mathbf{h}}_{i}^{t} &= \text{LSTM}_{\text{DEC}}(\hat{\mathbf{h}}_{i}^{t-1}, \hat{\mathbf{e}}_{i}^{t};\mathbf{W}_{\text{DEC}}),  \label{eq:lstm_dec} \\
    (\hat{x}_{i}^{t}, \hat{y}_{i}^{t}) &= \mathbf{W}_\text{o}\hat{\mathbf{h}}_{i}^{t} + \mathbf{b}_\text{o}, \label{eq:out}
\end{align}
where $\mathbf{W}_{\text{DEC}}$, $\mathbf{W}_\text{o}$, and $\mathbf{b}_\text{o}$ are parameters of the trajectory decoder $\text{LSTM}_{\text{DEC}}$ and the output layer.
Using the sampled scene gated social graph features, the trajectory decoder is able to generate multiple socially and environmentally acceptable trajectories.
To be more specific, given one sampled latent variable $\mathbf{z}_i$ in Eq.~\eqref{eq:sample}, one scene gated social graph feature $\mathbf{\hat{g}}_{i}^{\ol}$ will be decoded by Eq.~\eqref{eq:vae_decoder} and then predictions will be decoded through the trajectory decoder recursively (Eqs.~\eqref{eq:dec_init}-\eqref{eq:out}).

\subsection{Implementation Details}

We model the embedding layer $\phi$ in Eq.~\eqref{eq:emb} as a single layer perceptron which outputs 32-dimensional embedding vectors for the input location coordinates.
The dimensions of the hidden states of the LSTM layers for both the trajectory and social graph encoders are also 32.
We use a single graph convolutional layer in the GCN to encode each social graph.
The hidden dimension of the trajectory LSTM decoder is set to 64.

For the scene encoder module, the input video image is firstly resized to $224\times224$.
Similar to SoPhie~\cite{Sadeghian_2019_CVPR}, the encoded semantic feature of 256 channels is then passed through a convolutional layer and embedded to a scene feature $\mathbf{s}_i^{\ol} \in\mathds{R}^{32}$ through a single fully connected layer.
This fully connected layer acts like a transformation so that scene features learn to ``align'' with social graph features and are then used as gates on these features. 
This overall operation abstracts the redundant spatial information in scene images to features that are essential for trajectory prediction.
As the element-wise product operation is performed between $\mathbf{s}_i^{\ol}$ and the encoded social graph feature  $\mathbf{g}_{i}^{\ol}$ (see Eq.~\eqref{eq:social_scene}), both $\mathbf{g}_{i}^{\ol}$ and $\mathbf{G}_{i}^{\ol}$ are 32-dimensional also.
For the VAE module, $\bm{\mu}_i$, $\bm{\gamma}_i$, and the latent vector ${\bf z}_i$, for all $i$, are $\mathds{R}^8$ vectors. 

We use the Adam optimizer by minimizing the loss $\mathcal{L}_i$ given below for each $i^{\text{th}}$ POI:
$
    \mathcal{L}_i = \| {\bf{X}}_{\gt}^{i} - {\hat{\bf{X}}}_{\pred}^{i} \|_2^2 + D_{\text{KL}} (Q(\mathbf{z}_i | \mathbf{G}_{i}^{\ol}) ||P(\mathbf{z}_i)),  
$
where the L$_2$ loss in the first term represents the trajectory prediction loss
and the second term is the regularization loss which measures the Kullback-Leibler divergence (KLD) of the distribution $Q$ given in Eq.~\eqref{eq:sample} and the prior distribution $P(\mathbf{z}_i) := {\cal N}({\bf 0},I)$. 
Similar to~\cite{Lee_2017_CVPR,Li_2019_CVPR}, we adopt the reparameterization trick~\cite{kingma2013auto} in our experiments.
The learning rate and batch size are 0.001 and 128.

\begin{table*}[!tbp]
\centering
\caption{Quantitative results of various methods. The ADE / FDE values reported are in meters. The results with a $\dagger$ are taken from the corresponding papers. The results with a $\ddagger$ are taken from~\cite{Gupta_2018_CVPR}. Top performers for each column are marked in boldface.}
\footnotesize{
\addtolength{\tabcolsep}{-0.2ex}
\begin{tabular}{|lc|c|c|c|c|c||c|}
\hline
\multirow{2}{*}{Method} &
\multirow{2}{*}{\scriptsize{\texttt{\#}}} &
\multicolumn{6}{c|}{Scenes of the ETH \& UCY datasets} \\ \cline{3-8} 
 &  & ETH & HOTEL & UNIV & ZARA1 & ZARA2 & Average \\ \hline
Social-LSTM~\cite{Alahi_2016_CVPR}$^\ddagger$ $_{\textit{CVPR}16}$ & 1 & 1.09 / 2.35 & 0.79 / 1.76 & 0.67 / 1.40 & 0.47 / 1.00 & 0.56 / 1.17 & 0.72 / 1.54 \\ 
SGAN 1V-1~\cite{Gupta_2018_CVPR}$^\dagger$ $_{\textit{CVPR}18}$ & 1 & 1.13 / 2.21 & 1.01 / 2.18 & 0.60 / 1.28 & 0.42 / 0.91 & 0.52 / 1.11 & 0.74 / 1.54 \\ 
MX-LSTM~\cite{Hasan_2018_CVPR}$^\dagger$ $_{\textit{CVPR}18}$ & 1 & -- & -- & 0.49 / 1.12 & 0.59 / 1.31 & 0.35 / 0.79 & -- \\ 
Nikhil \& Morris~\cite{nikhil2018convolutional}$^\dagger$ $_{\textit{ECCV}18}$ &1 & 1.04 / 2.07 & 0.59 / 1.17 & 0.57 / 1.21 & 0.43 / 0.90 & 0.34 / 0.75 & 0.59 / 1.22 \\ 
Liang~\etal~\cite{Liang_2019_CVPR}$^\dagger$ $_{\textit{CVPR}19}$ & 1 & 0.88 / 1.98 & 0.36 / 0.74 & 0.62 / 1.32 & 0.42 / 0.90 & 0.34 / 0.75 & 0.52 / 1.14 \\ 
MATF~\cite{Zhao_2019_CVPR}$^\dagger$ $_{\textit{CVPR}19}$ & 1 & 1.33 / 2.49 & 0.51 / 0.95 & 0.56 / 1.19 & 0.44 / 0.93 & 0.34 / 0.73 & 0.64 / 1.26 \\ 
SAGCN~\cite{sun2019socially}$^\dagger$ $_{\textit{ITNEC}19}$ & 1 & 0.90 / 1.96 & 0.41 / 0.83 & 0.57 / 1.19 & {0.41} / 0.89 & 0.32 / 0.70 & 0.52 / 1.11 \\ 
SR-LSTM\_1~\cite{zhang2019sr}$^\dagger$ $_{\textit{CVPR}19}$ & 1 & 0.64 / 1.28 & 0.39 / 0.78 & 0.52 / 1.13 & 0.42 / 0.92 & 0.34 / 0.74 & 0.46 / 0.97 \\ 
SR-LSTM\_2~\cite{zhang2019sr}$^\dagger$ $_{\textit{CVPR}19}$ & 1 & 0.63 / 1.25 & 0.37 / 0.74 & \textbf{0.51} / \textbf{1.10} & {0.41} / 0.90 & 0.32 / 0.70 & \textbf{0.45} / 0.94 \\ 
STGAT 1V-1~\cite{huang2019stgat}$^\dagger$ $_{\textit{ICCV}19}$ & 1 & 0.88 / 1.66 & 0.56 / 1.15 & 0.52 / 1.13 & {0.41} / 0.91 & {0.31} / {0.68} & 0.54 / 1.11 \\ 
RSBG~\cite{Sun_2020_CVPR}$^\dagger$ $_{\textit{CVPR}20}$ & 1 & 0.80 / 1.53 & 0.33 / 0.64 & 0.59 / 1.25 & \textbf{0.40} / \textbf{0.86} & \textbf{0.30} / \textbf{0.65} & 0.48 / 0.99 \\ \hline
\name\ (ours) & 1 & \textbf{0.62} / \textbf{1.23} & \textbf{0.29} / \textbf{0.54} & 0.62 / 1.27 & {0.41} / {0.87}  & 0.33 / 0.70  & \textbf{0.45} / \textbf{0.92} \\ \hline \hline
CVAE~\cite{zhang2019stochastic}$^\dagger $ $_{\textit{arXiv}19}$& 20 & 0.93 / 1.94 & 0.52 / 1.03 & 0.59 / 1.27 & 0.41/ 0.86 & 0.33 / 0.72 & 0.53 / 1.11 \\ 
SGAN 20V-20~\cite{Gupta_2018_CVPR}$^\dagger$ $_{\textit{CVPR}18}$& 20 & 0.81 / 1.52 & 0.72 / 1.61 & 0.60 / 1.26 & 0.34 / 0.69 & 0.42 / 0.84 & 0.58 / 1.18 \\ 
SoPhie~\cite{Sadeghian_2019_CVPR}$^\dagger$ $_{\textit{CVPR}19}$& 20 & 0.70 / 1.43 & 0.76 / 1.67 & 0.54 / 1.24 & 0.30 / 0.63 & 0.38 / 0.78 & 0.54 / 1.15 \\ 
Liang~\etal~\cite{Liang_2019_CVPR}$^\dagger$ $_{\textit{CVPR}19}$& 20 & 0.73 / 1.65 & 0.30 / 0.59 & 0.60 / 1.27 & 0.38 / 0.81 & 0.31 / 0.68 & 0.46 / 1.00 \\ 
Social Ways~\cite{Amirian_2019_CVPR_Workshops}$^\dagger$ $_{\textit{CVPR}19}$& 20 & \textbf{0.39} / \textbf{0.64} & 0.39 / 0.66 & 0.55 / 1.31 & 0.44 / 0.64 & 0.51 / 0.92 & 0.46 / 0.83 \\ 
MATF GAN~\cite{Zhao_2019_CVPR}$^\dagger$ $_{\textit{CVPR}19}$& 20 & 1.01 / 1.75 & 0.43 / 0.80 & \textbf{0.44} / {0.91} & 0.26 / \textbf{0.45} & 0.26 / 0.57 & 0.48 / 0.90 \\ 
IDL~\cite{Li_2019_CVPR}$^\dagger$ $_{\textit{CVPR}19}$& 20 & 0.59 / 1.30 & 0.46 / 0.83 & 0.51 / 1.27 & \textbf{0.22} / {0.49} & \textbf{0.23} / {0.55} & \textbf{0.40} / 0.89 \\ 
Zhang~\etal~\cite{zhang2019stochastic}$^\dagger$ $_{\textit{arXiv}19}$& 20 & 0.75 / 1.63 & 0.63 / 1.01 & 0.48 / 1.08 & 0.30 / 0.65 & 0.26 / 0.57 & 0.48 / 0.99 \\ 
STGAT 20V-20~\cite{huang2019stgat}$^\dagger$ $_{\textit{ICCV}19}$& 20 & 0.65 / 1.12 & 0.35 / 0.66 & 0.52 / 1.10 & 0.34 / 0.69 & 0.29 / 0.60 & 0.43 / 0.83 \\ 
Social-BiGAT~\cite{socialbigat_neurips19}$^\dagger$ $_{\textit{NeurIPS}19}$& 20 & 0.69 / 1.29 & 0.49 / 1.01 & 0.55 / 1.32 & 0.30 / 0.62 & 0.36 / 0.75 & 0.48 / 1.00 \\ 
{Social-STGCNN}~\cite{Mohamed_2020_CVPR}$^\dagger$ $_{\textit{CVPR}20}$& 20 & 0.64 / 1.11 & 0.49 / 0.85 & \textbf{0.44} / \textbf{0.79} & 0.34 / 0.53 & 0.30 / \textbf{0.48} & 0.44 / \textbf{0.75} \\
{PMP-NMMP}~\cite{Hu_2020_CVPR}$^\dagger$ $_{\textit{CVPR}20}$& 20 & 0.61 / 1.08 & 0.33 / 0.63 & 0.53 / 1.17 & 0.28 / 0.61 & 0.28 / 0.59 & 0.41 / 0.82 \\
{Sun~\etal}~\cite{Sun1_2020_CVPR}$^\dagger$ $_{\textit{CVPR}20}$& 20 & 0.69 / 1.24 & 0.43 / 0.87 & 0.52 / 1.11 & 0.32 / 0.66 & 0.29 / 0.61 & 0.44 / 0.90 \\ \hline
\name\ (ours)& 20 & 0.54 / 1.07 & \textbf{0.24} / \textbf{0.45} & 0.57 / 1.19 & 0.35 / 0.79 & 0.28 / 0.59 & \textbf{0.40} / {0.82} \\ \hline \hline
Trajectron~\cite{Ivanovic_2019_ICCV}$^\dagger$ $_{\textit{ICCV}19}$& 100 & \textbf{0.37} / \textbf{0.72} & 0.20 / \textbf{0.35} & 0.48 / 0.99 & 0.32 / 0.62 & 0.34 / 0.66 & 0.34 / 0.67 \\ \hline
\name\ (ours)& 100 & 0.44 / 0.81 & \textbf{0.19} / 0.36 & \textbf{0.38} / \textbf{0.81} & \textbf{0.29} / \textbf{0.59} & \textbf{0.24} / \textbf{0.53} & \textbf{0.31} / \textbf{0.62} \\ \hline
\end{tabular}}
\label{tab:all_results}
\vspace{-2ex}
\end{table*}
\section{Experiments}
\label{sec:4}
\subsection{Datasets, Metrics, and Preprocessing}

\noindent \textbf{Datasets.}
As we focus on the prediction of pedestrian trajectories in this paper, we evaluate our method using the two public
ETH~\cite{pellegrini2009you} and UCY~\cite{lerner2007crowds} datasets 
that are widely adopted in the literature.
There are 5 scenes, commonly referred to as ETH, HOTEL, 
UNIV, ZARA1 ,and ZARA2, 
with a total of 1535 different trajectories.
The location coordinates of these trajectories are labeled in meters.
Following the previous work of others~\cite{Alahi_2016_CVPR,Gupta_2018_CVPR,Sadeghian_2019_CVPR,zhang2019sr,huang2019stgat}, the leave-one-out experiment approach is used, \ie, 4 scenes are used for training and the remaining scene for testing.
The observation period and prediction period are 3.2 seconds ($\ol=8$ frames) and 4.8 seconds ($\pl=12$ frames).

\smallskip

\noindent \textbf{Metrics.}
Similar to the previous work~\cite{Alahi_2016_CVPR,Gupta_2018_CVPR,Hasan_2018_CVPR,xue2019location}, we use two metrics: the average displacement error (ADE)~\cite{pellegrini2009you} and the final displacement error (FDE)~\cite{Alahi_2016_CVPR} to quantitatively evaluate the trajectory prediction performance of each method.
Smaller values of these metrics indicate better performance.

\smallskip

\noindent \textbf{Trajectory Preprocessing and Augmentation.}
We normalized all the trajectories so that their location coordinates are in the range $[-1,1]$. After prediction, the inverse normalization is applied to yield the ADE and FDE values back in meters.
Similar to~\cite{Sadeghian_2019_CVPR,zhang2019sr}, we also performed data augmentation.
Specifically, we applied the following operations: (i)~a sliding time window of stride 1 is used to convert long trajectories into multiple trajectories of length $\ol + \pl$. (ii)~extra trajectories are generated by rotating all the trajectories and the corresponding video images by $90^\circ$.

\subsection{Comparison with Existing Methods}

We compare the performance of \name\ against 20 existing methods listed below:
\textit{Social-LSTM}~\cite{Alahi_2016_CVPR},
\textit{CVAE}~\cite{zhang2019stochastic},
\textit{SGAN}~\cite{Gupta_2018_CVPR},
\textit{MX-LSTM}~\cite{Hasan_2018_CVPR},
\textit{Nikhil and Morris}~\cite{nikhil2018convolutional},
\textit{Liang~\etal}~\cite{Liang_2019_CVPR},
\textit{SoPhie}~\cite{Sadeghian_2019_CVPR},
\textit{MATF}~\cite{Zhao_2019_CVPR},
\textit{Social Ways}~\cite{Amirian_2019_CVPR_Workshops},
\textit{IDL}~\cite{Li_2019_CVPR},
\textit{Zhang~\etal}~\cite{zhang2019stochastic},
\textit{SAGCN}~\cite{sun2019socially}, 
\textit{SR-LSTM}~\cite{zhang2019sr},
\textit{STGAT}~\cite{huang2019stgat}, 
\textit{Social-BiGAT}~\cite{socialbigat_neurips19},
\textit{Trajectron}~\cite{Ivanovic_2019_ICCV},
\textit{Sun~\etal}~\cite{Sun1_2020_CVPR},
\textit{PMP-NMMP}~\cite{Hu_2020_CVPR},
\textit{Social-STGCNN}~\cite{Mohamed_2020_CVPR}, and
\textit{RSBG}~\cite{Sun_2020_CVPR}.
For the SR-LSTM, two configurations are reported in~\cite{zhang2019sr}: SR-LSTM\_1 with a 2 meters neighborhood size, and SR-LSTM\_2 with a 10 meters neighborhood size.
For the MX-LSTM, only evaluation results on the UCY dataset are given in~\cite{Hasan_2018_CVPR}.
As the results of DESIRE~\cite{Lee_2017_CVPR} on the ETH and UCY datasets are not reported in the authors' paper, we include only the results of \textit{CVAE} (reported in~\cite{zhang2019stochastic}) in Table~\ref{tab:all_results}.
It should be noted that, although CVAE was introduced and used to generate multimodal trajectories in DESIRE~\cite{Lee_2017_CVPR},
the implementation in~\cite{zhang2019stochastic} does not include the refinement module of DESIRE.

Table~\ref{tab:all_results} shows the ADEs and FDEs in meters of these pedestrian trajectory prediction methods.
We separate these methods and their variants based on whether they generate only one prediction (upper half of the table, $\texttt{\#}\!=\!1$) or multiple predictions (bottom half, $\texttt{\#}\!=\!20$ or $\texttt{\#}\!=\!100$) per observed trajectory.
The `\texttt{\#}' column indicates how many predictions are generated for each input observed trajectory.
As shown in the table, \name\ is able to achieve comparable results with the state-of-the-art methods for the single prediction scenario.
Both \name\ and SR-LSTM\_2 attain the smallest average ADE of 0.45m. Moreover, \name\ takes the lead for the FDE metric at 0.92m.  

\begin{table*}[!tbp]
\centering
\caption{The ADE / FDE values in meters of the variants of \name. The top-performing single prediction and multimodal prediction variants for each scene and on average are marked in boldface.}
\footnotesize{
\begin{tabular}{|l|ccc|c|c|c|c|c|c|c||c|}
\hline
\multirow{2}{*}{Variant} & \multicolumn{3}{c|}{Modules} & \multirow{2}{*}{Merge} & \multirow{2}{*}{\scriptsize{\texttt{\#}}}& \multicolumn{6}{c|}{Scenes of the ETH \& UCY datasets} \\ \cline{2-4}\cline{7-12}
 & SG & VAE & Scene & & & ETH & HOTEL & UNIV & ZARA1 & ZARA2 & Average \\ \hline
\name-v1 & \checkmark & \xmark & \xmark & N/A & 1 & 0.67 / 1.29 & 0.35 / 0.65 & 0.67 / 1.36 & 0.43 / 0.90 & 0.38 / 0.77 & 0.50 / 0.99 \\ \hline
\name-v2 & \xmark & \xmark & \checkmark & N/A & 1 & 0.76 / 1.58 & 0.40 / 0.75 & 0.73 / 1.40 & 0.45 / 0.95 & 0.42 / 0.90 & 0.55 / 1.12 \\ \hline
\multirow{2}{*}{\name-v3} & \xmark & \checkmark & \checkmark & N/A & 1 & 0.79 / 1.61 & 0.42 / 0.79 & 0.70 / 1.41 & 0.46 / 0.98 & 0.41 / 0.87 & 0.56 / 1.13 \\
\cline{2-12} 
 & \xmark & \checkmark & \checkmark & N/A & 20 & 0.73 / 1.48 & 0.39 / 0.76 & 0.66 / 1.36 & 0.42 / 0.92 & 0.36 / 0.79 & 0.51 / 1.06 \\ \hline
\multirow{2}{*}{\name-v4} & \checkmark & \checkmark & \xmark & N/A & 1 & 0.66 / 1.28 & 0.34 / 0.61 & 0.68 / 1.34 & 0.43 / 0.89 & 0.38 / 0.76 & 0.50 / 0.98 \\ 
\cline{2-12} 
 & \checkmark & \checkmark & \xmark & N/A & 20 & 0.62 / 1.13 & 0.30 / 0.56 & 0.61 / 1.28 & 0.41 / 0.85 & 0.34 / 0.69 & 0.46 / 0.90 \\ \hline
 \name-v5 & \checkmark & \xmark & \checkmark & gating & 1 & \textbf{0.62} / \textbf{1.21} & \textbf{0.29} / 0.55 &  \textbf{0.61} / \textbf{1.26} & 0.42 / 0.88 & \textbf{0.33} / 0.73 & \textbf{0.45} / 0.93 \\ \hline
\multirow{2}{*}{\name-$\alpha$} & \checkmark & \checkmark & \checkmark & add & 1 & 0.69 / 1.25 & 0.31 / 0.56 & 0.64 / 1.31 & 0.44 / 0.93 & 0.37 / 0.76 & 0.49 / 0.96 \\ 
\cline{2-12} 
 & \checkmark & \checkmark & \checkmark & add & 20 & 0.65 / 1.19 & 0.28 / 0.51 & 0.60 / 1.25 & 0.38 / 0.83 & 0.31 / 0.67 & 0.44 / 0.89 \\ \hline
\multirow{2}{*}{\name-$\beta$} & \checkmark & \checkmark & \checkmark & concat & 1 & 0.64 / 1.22 & 0.30 / 0.56 & 0.64 / 1.30 & 0.43 / 0.93 & 0.36 / 0.76 & 0.47 / 0.95 \\ 
\cline{2-12} 
 & \checkmark & \checkmark & \checkmark & concat & 20 & 0.59 / 1.18 & 0.28 / 0.50 & 0.59 / 1.23 & 0.39 / 0.83 & 0.30 / 0.63 & 0.43 / 0.87 \\ \hline
\multirow{2}{*}{\name} & \checkmark & \checkmark & \checkmark & gating & 1 & \textbf{0.62} / {1.23} & \textbf{0.29} / \textbf{0.54} & 0.62 / 1.27 & \textbf{0.41} / \textbf{0.87}  & \textbf{0.33} / \textbf{0.70}  & \textbf{0.45} / \textbf{0.92} \\
\cline{2-12} 
 & \checkmark & \checkmark & \checkmark & gating & 20 & \textbf{0.54} / \textbf{1.07} & \textbf{0.24} / \textbf{0.45} & \textbf{0.57} / \textbf{1.19} & \textbf{0.35} / \textbf{0.79} & \textbf{0.28} / \textbf{0.59} & \textbf{0.40} / \textbf{0.82} \\ \hline
\end{tabular}}
\label{tab:ablation}
\end{table*}

When 20 predicted trajectories are generated (the lower half of Table~\ref{tab:all_results}), we observe that our \name\ also achieves the second smallest FDE of 0.82m and the smallest average ADE of 0.40m (the latter is on par with IDL).
Top performers in the lower half of the table include Social Ways, IDL, MATF GAN, Social-STGCNN, and our \name, each of which achieves the smallest ADE/FDE for one or more scenes.
For the four methods that appear in both halves of the table, namely SGAN, MATF, STGAT, Liang~\etal, and our \name, we observe that their average percentage reductions of ADEs and FDEs for multimodal predictions are large (or small) when their ADEs and FDEs for single predictions are large (or small). Thus, compared to the other three methods, our \name, which already manifests good performance for single predictions, only has a smaller drop (around 11\%) in ADEs and FDEs for multimodal predictions.
When more predictions are generated ($\texttt{\#}\!=\!100$), our \name\ outperforms the state-of-the-art Trajectron as well.
Overall, these results demonstrate the high accuracy of our proposed method in forecasting  single and multimodal future trajectories.

\subsection{Ablation Study}
To fully study the effectiveness of each encoder of our proposed method, five different variants, labeled as \name-v1 to \name-v5 in Table~\ref{tab:ablation},  with one or more of the social graph (SG) encoder, scene encoder, and the VAE module disabled are analyzed on the ETH \& UCY datasets.
A tick under the \textit{Modules} column denotes that the module is enabled; a cross denotes that it is disabled. With the VAE encoder/decoder enabled, the \name-v3 and \name-v4 variants can forecast either one or 20 predictions, as shown under the `\texttt{\#}' column.
For the variants \name-v1, \name-v2, and \name-v5 that do not include the VAE module, the loss function only contains the L$_2$ prediction loss term in the training phase.
For all of the above five variants, the scene gated social graph feature ${\mathbf{G}_{i}^{\ol}}$ given in Eq.~\eqref{eq:social_scene} reduces to $\mathbf{g}_{i}^{\ol}$ if the scene encoder is disabled, and to ${\mathbf{s}_{i}^{\ol}}$ if the social graph encoder is disabled.

To further validate the effectiveness of the proposed gating operation on the scene gated social graph features, we design two more variants which implement different methods of merging the social graph and scene features:
\name-$\alpha$, which element-wise adds these features; and
\name-$\beta$, which directly concatenates them.
As these variants are designed to single out the gating operation for evaluation, they have all the modules enabled like \name.

From the results in Table~\ref{tab:ablation}, we observe that
\name\ outperforms all the variants. This indicates that it is important to include all the three modules in \name.
What is also noticeable from the table is, while including the scene influence into \name\ helps to improve the prediction performance, it can not work well alone without the social graph encoder. This is evident from the poorer performance of \name-v2 and \name-v3 when the scene encoder is enabled but the SG encoder is disabled. 
On the other hand, when the SG encoder is enabled, as in \name-v1 and \name-v4, lower ADEs and FDEs are achieved.
It is interesting that the models with VAE and the same without VAE produce very similar results under the single prediction case. For example,
\name-v4 ($\texttt{\#}\!=\!1$) and \name-v1 have similar performance; so do \name-v3 ($\texttt{\#}\!=\!1$) and \name-v2; likewise for \name ($\texttt{\#}\!=\!1$) and \name-v5.
These results indicate that the VAE module performs well in the gated feature reconstruction process.
Compared to the first four variants which have either the scene feature or the social graph feature removed, both \name-$\alpha$ and \name-$\beta$ achieve better performance in most scenes and on average.
This further demonstrates the importance of incorporating both social and scene cues.
For different merging methods of the social graph and scene features, we observe that the concatenation operation (\name-$\alpha$) performs slightly better than simple addition (\name-$\beta$).
However, comparing with \name\ where the proposed gating operation is implemented, both \name-$\alpha$ and \name-$\beta$ clearly have inferior performance. This performance comparison  demonstrates the superiority of our scene gated social graph features proposed in \name.

\begin{figure*}[!tbp]
  \centering
  \subfigure{
  \includegraphics[trim=50 35.0 0 45, clip,width=3.2cm]{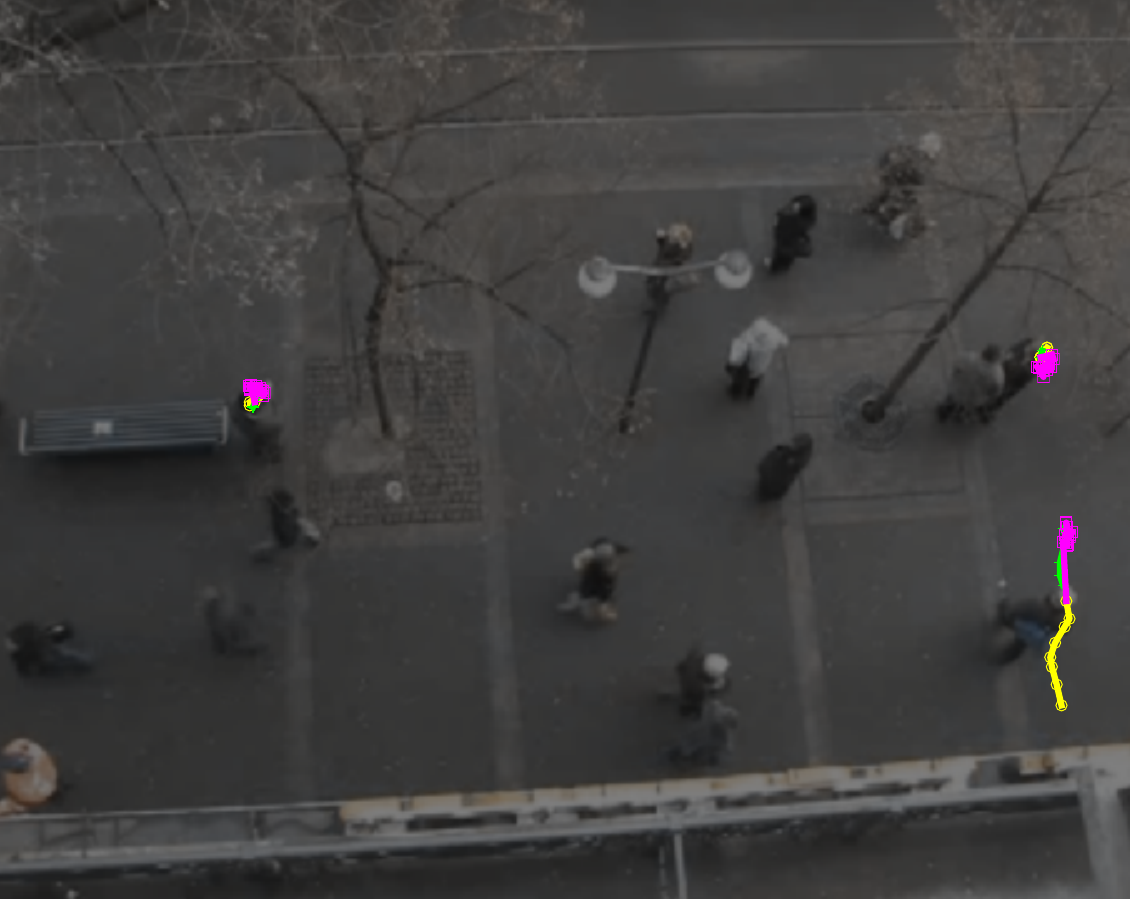}}
  \subfigure{
  \includegraphics[trim=52 40.0 20 40, clip,width=3.2cm]{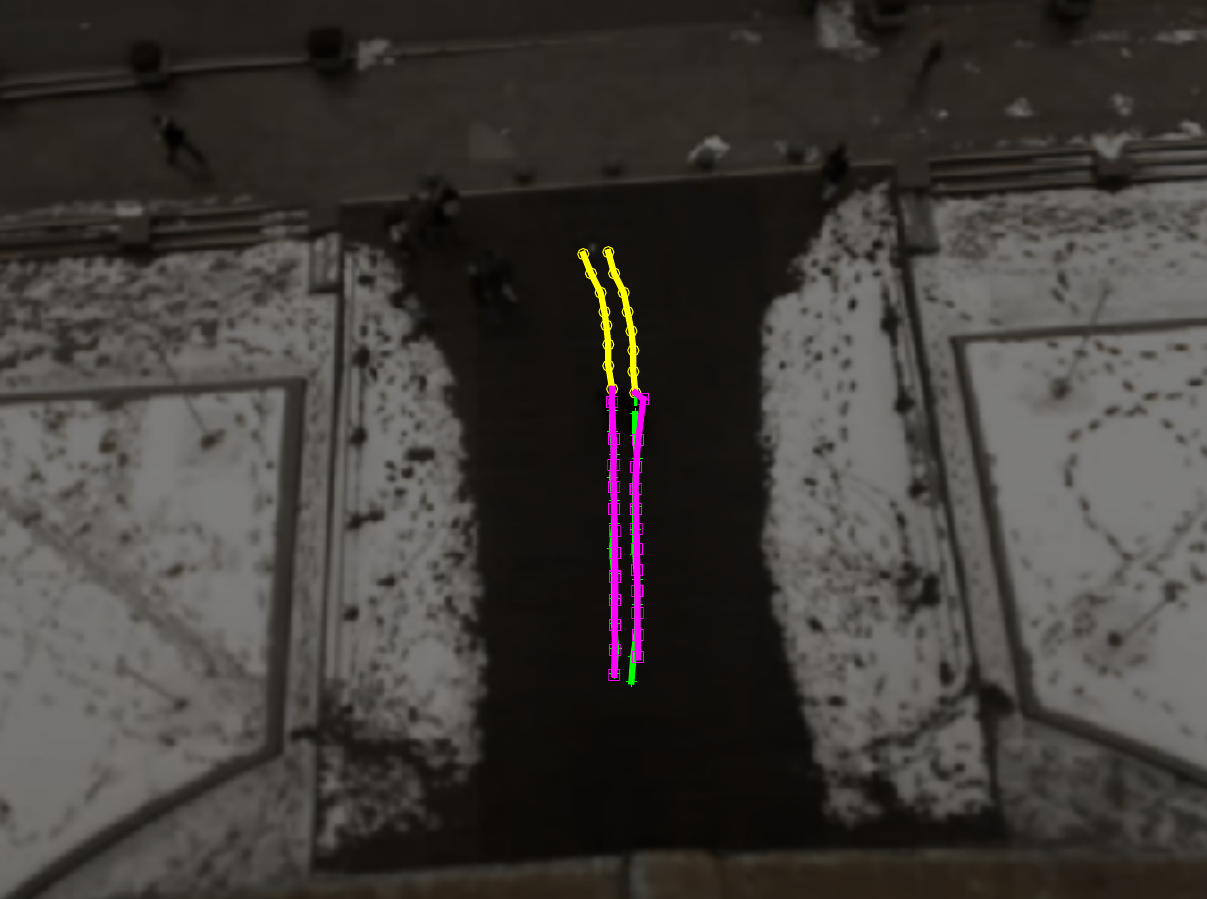}}
  \subfigure{
  \includegraphics[trim=40 80.0 10 0, clip,width=3.2cm]{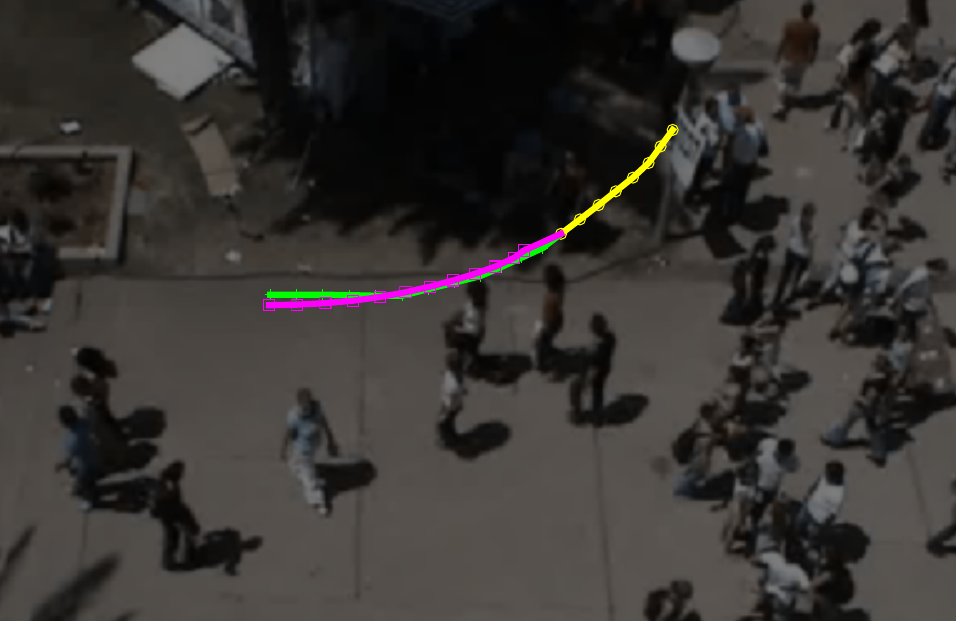}} 
  \subfigure{
  \includegraphics[trim=5 10.0 15 50, clip,width=3.2cm]{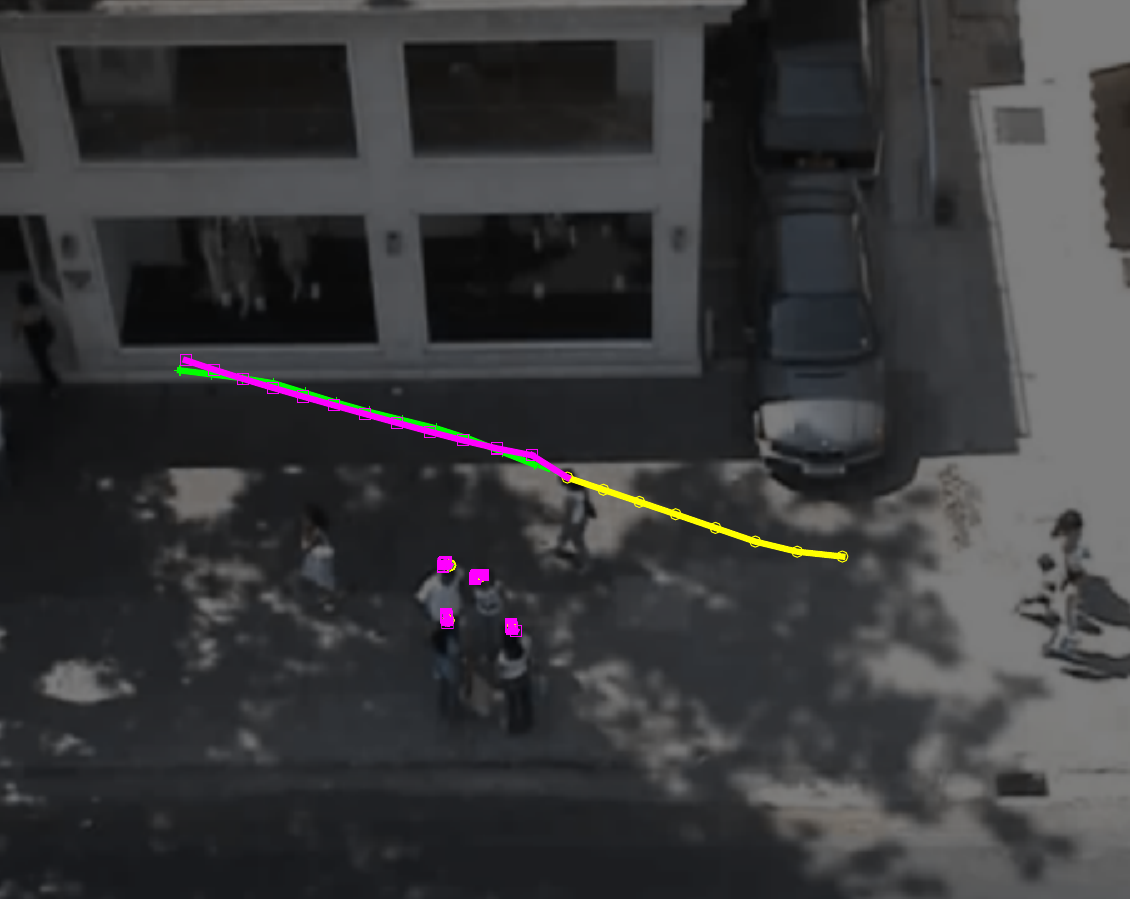}}
    \subfigure{
  \includegraphics[trim=0 65.0 150 80, clip,width=3.2cm]{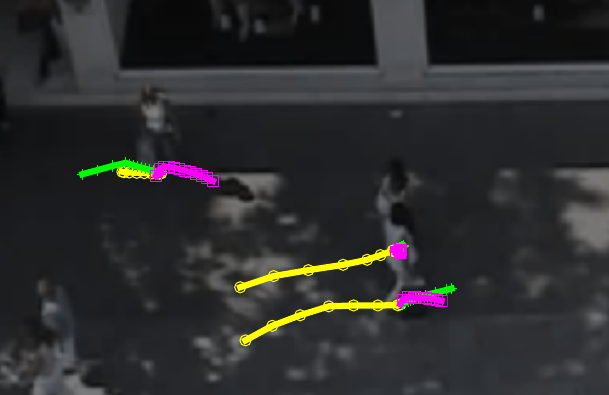}} 
  \\ [-4ex]
  \parbox{3.2cm}{\textcolor{white}{{~~\small (a) }}} \,
  \parbox{3.2cm}{\textcolor{white}{{~~\small (b) }}} \,
  \parbox{3.2cm}{\textcolor{white}{{~~\small (c) }}} \,
  \parbox{3.2cm}{\textcolor{white}{{~~\small (d) }}} \,
  \parbox{3.2cm}{\textcolor{white}{{~\small (e) }}} \\ \vspace{-1ex}
  \subfigure{
  \includegraphics[trim=50 35.0 0 45, clip,width=3.2cm]{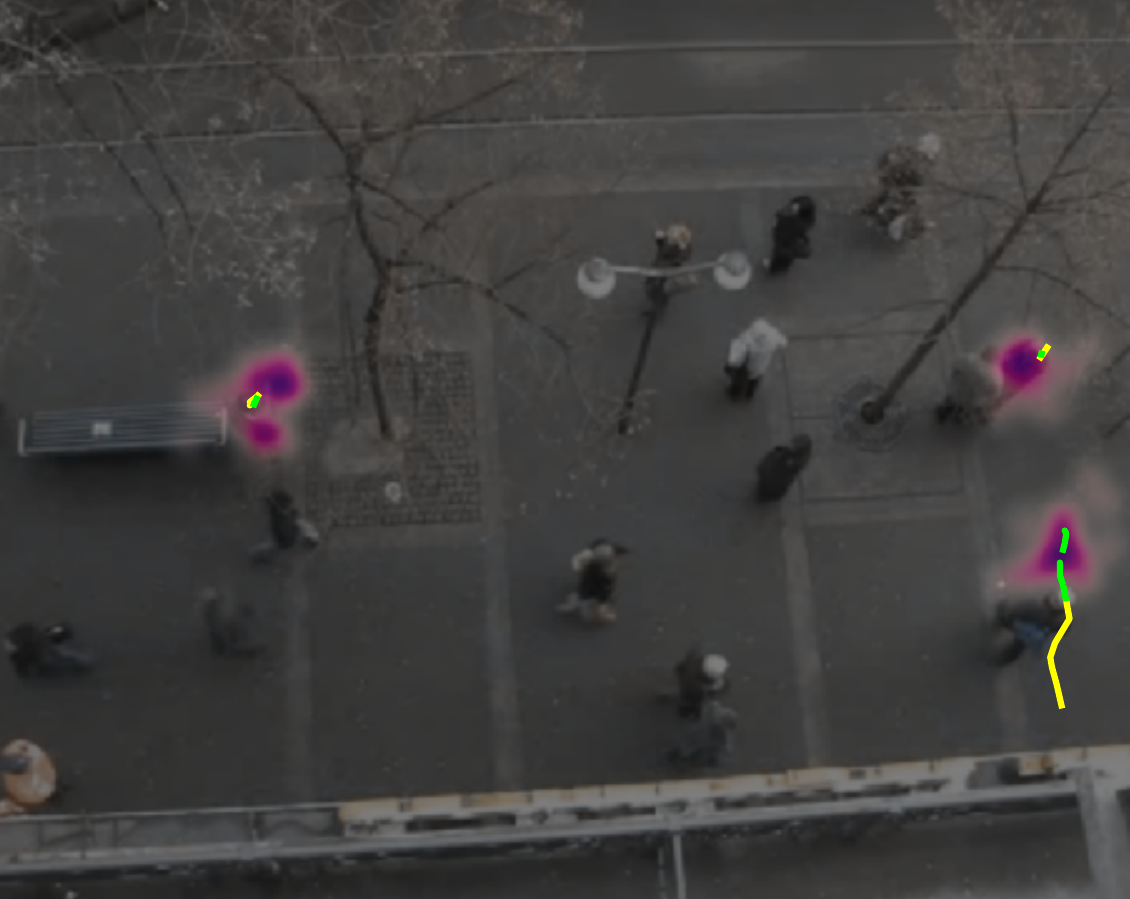}}
  \subfigure{
  \includegraphics[trim=52 40.0 20 40, clip,width=3.2cm]{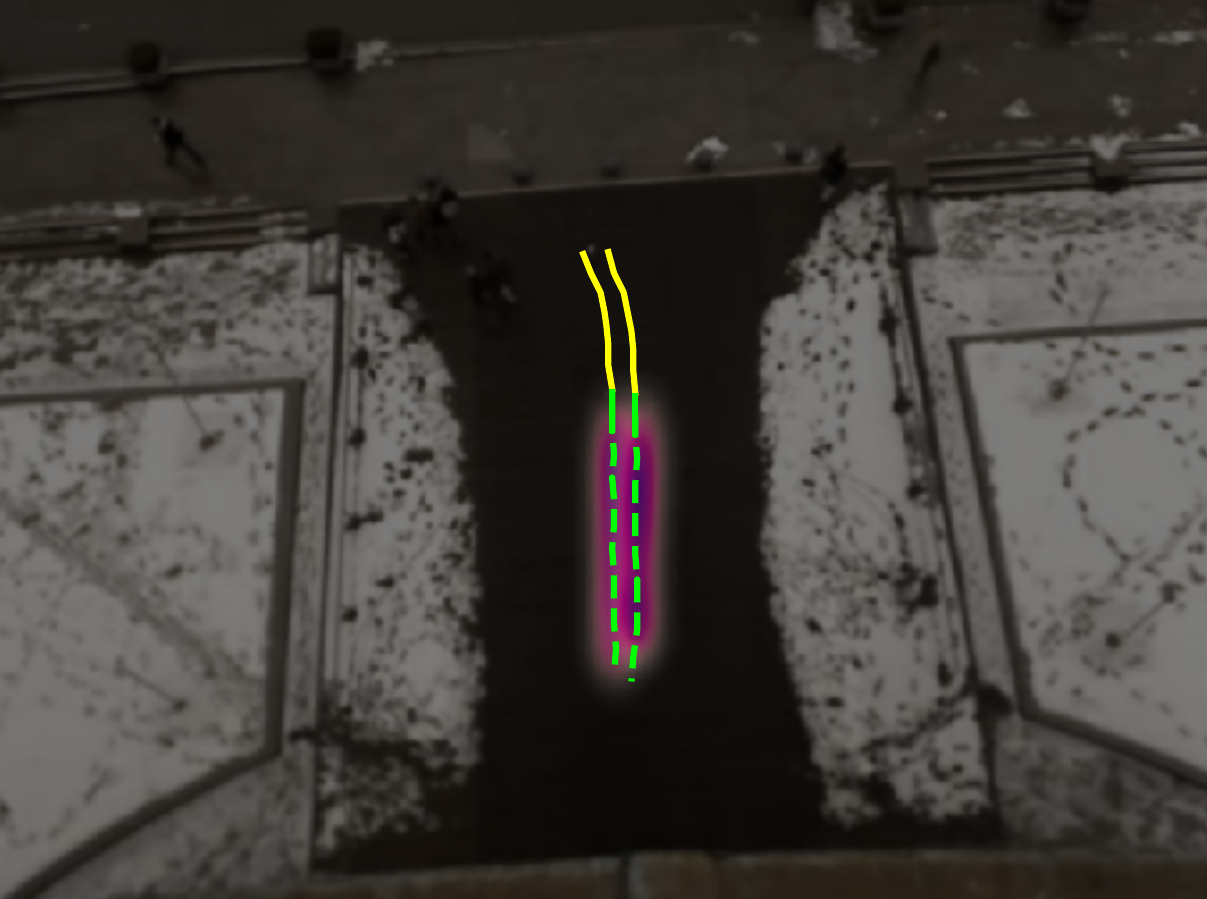}}
  \subfigure{
  \includegraphics[trim=40 80.0 10 0, clip,width=3.2cm]{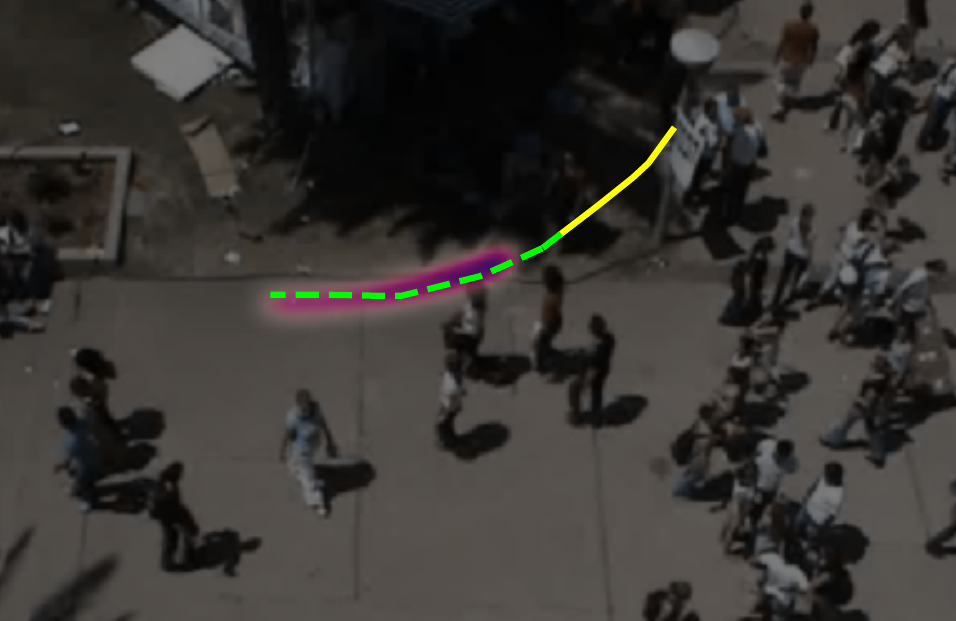}}
\subfigure{
  \includegraphics[trim=5 10.0 15 50, clip,width=3.2cm]{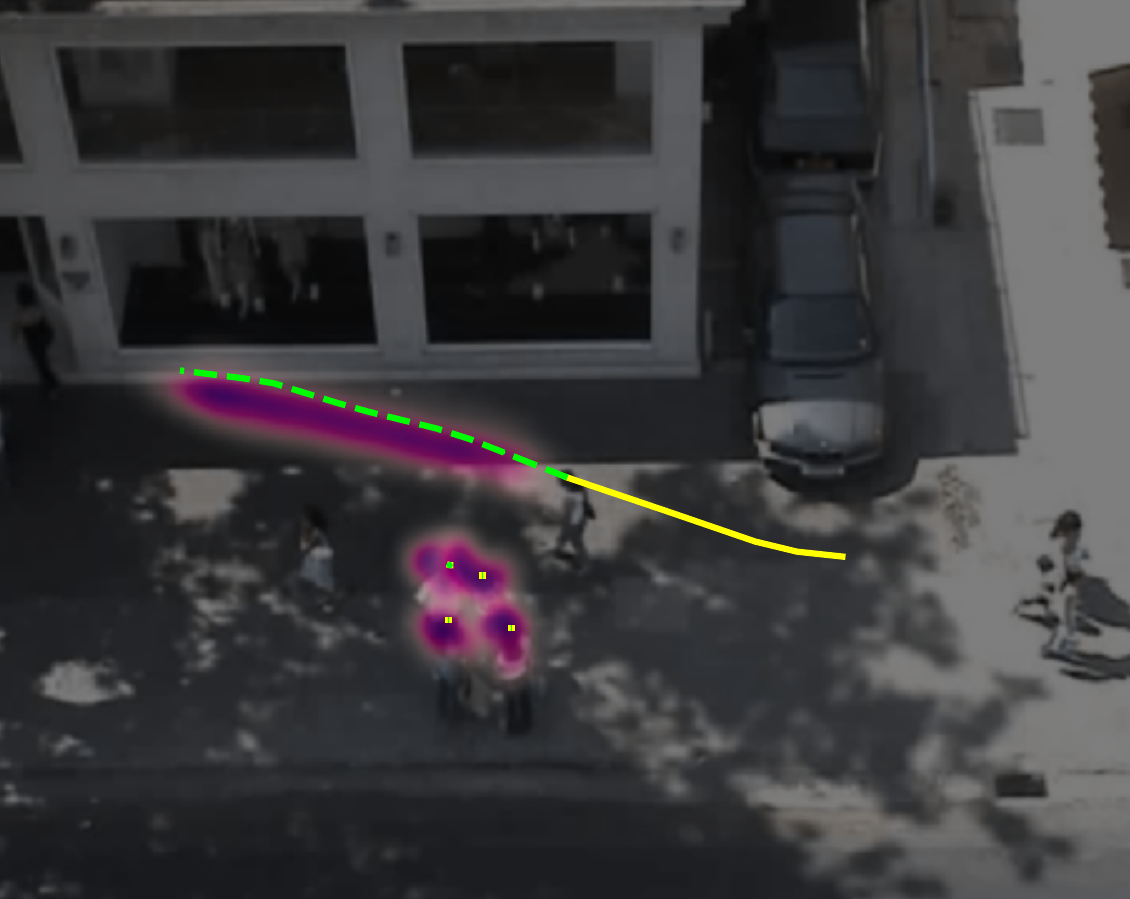}}
\subfigure{
  \includegraphics[trim=0 65.0 150 80, clip,width=3.2cm]{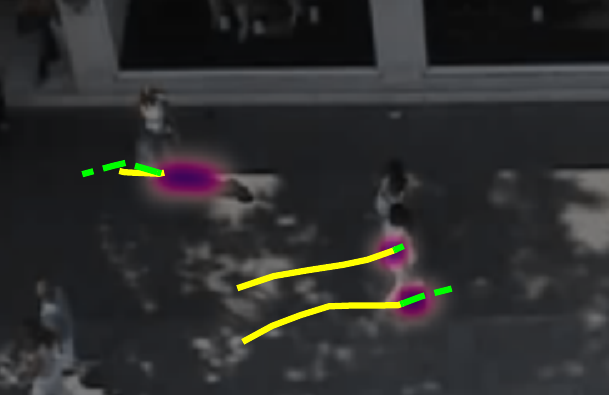}}
    \\ [-4ex]
  \parbox{3.2cm}{\textcolor{white}{~~{\small (f)}}} \,
  \parbox{3.2cm}{\textcolor{white}{~~{\small (g)}}} \,
  \parbox{3.2cm}{\textcolor{white}{~~{\small (h)}}} \,
  \parbox{3.2cm}{\textcolor{white}{~~{\small (i)}}} \,
  \parbox{3.2cm}{\textcolor{white}{~{\small (j)}}} \\\vspace{2ex}
\caption{Examples of predictions generated by~\name\ on the ETH and UCY scenes. Legend: yellow: input observed trajectories; green: ground truth trajectories; pink: generated predictions.
All 20 generated trajectories for each POI are shown as heatmaps in the second row.
Failure cases are given in the last column((e) \& (j)).}
\label{fig:pred}
\end{figure*}

\subsection{Memory Usage}

\begin{table}
\addtolength{\tabcolsep}{-0.15ex}
\centering
\caption{The GPU memory usage in MB.}
\scriptsize{
\begin{tabular}{|l|c|c|c|c|}
\cline{1-5}
\multicolumn{1}{|c|}{} & STGAT & \mbox{\name-v3} & \mbox{\name-v4} & \name \\ \hline
Training & 7503 & 1577 & 505 & 1597 \\ 
Evaluation & 593 & 499 & 457 & 509 \\ \hline
\end{tabular}}
\label{tab:memory}
\end{table}

In Table~\ref{tab:memory}, we compare the GPU memory usage of the state-of-the-art STGAT~\cite{huang2019stgat} against \name.
The values of STGAT are directly taken from~\cite{huang2019stgat} and, for a fair comparison, we use the same batch size setting as in~\cite{huang2019stgat}.
STGAT's GPU memory usage for the training phase is 4.7 times and for the evaluation phase is 1.17 times of our \name's. 
The reason for the large memory usage of STGAT is the use of the complete graph structure to represent the social relationship as oppose to the star graph structure used in our \name. 
We also report the GPU memory usage of \name-v3 and \name-v4 in Table~\ref{tab:memory}. 
While both variants have the VAE module enabled and can produce multimodal predictions, \name-v3 demands much more GPU memory than \name-v4 in order to process the semantic segmentation based scene encoder. Its GPU memory usage is almost the same as the full version \name. Thus, in the case where the GPU memory is limited, \name-v4 can be used instead of \name\ with a small compromise in prediction accuracy.

\subsection{Qualitative Results}

Some prediction results from our \name\ for different movement scenarios are shown in Fig.~\ref{fig:pred}. More examples are shown in the \textit{supplementary materials} document.
In the first row, the best trajectory of the 20 predictions is shown 
for each observed trajectory.
The background images have been darkened and blurred for better visualization.
We can see that \name\ is able to generate realistic predictions for different cases such as stopping (Fig.~\ref{fig:pred} (a)), walking together (Fig.~\ref{fig:pred} (b)), and turning (Fig.~\ref{fig:pred} (c)).
Figure~\ref{fig:pred} (c) shows that the predicted trajectories by \name\ are compatible with the scene layouts as \name\ incorporates the scene information through the scene encoder.
In addition, when pedestrians are walking together and stopping to chat with each other (Fig.~\ref{fig:pred} (d)), \name\ 
performs well and the generated trajectories are both socially and environmentally acceptable.
In the second row, all 20 prediction of each POI given in the first row are shown as heatmaps.
Taken the left POI in Fig.~\ref{fig:pred} (f) as an example, it is a stopping case. In the prediction phase, the pedestrian can remain still or resume walking in any direction.
The predicted heatmap shows a good coverage of this situation.
Moreover, we can see that the heatmap indicates that the chance of this pedestrian walking towards left is very low as he/she should avoid the obstacle (the bench on the left).
These visualizations demonstrate the ability of \name\ to model and predict pedestrians moving in a group (such as Fig.~\ref{fig:pred} (d) and (i)).

The last column of Fig.~\ref{fig:pred} shows some failure cases that \name\ currently can not handle well. 
For the top left POI of Fig.~\ref{fig:pred} (e), the POI is moving to the right (yellow trajectory) in the observation period but makes a sudden $180^\circ$ turn (green) shortly after being in the prediction period. 
As the multiple predictions shown in the corresponding heatmap in Fig.~\ref{fig:pred} (j) all forecast a right-ward movement, the most reasonable predicted trajectory in this case is the continuation of the observed trajectory (pink). 
As there are neither scene obstacles nor pedestrians nearby, no social or scene context information is available to help overcome this sudden U-turn scenario.
For the two POIs in the bottom of Fig.~\ref{fig:pred} (e), they are walking quite fast together initially but decelerating near the end of the observation period. 
One of the pedestrians almost comes to a stop in the prediction period.
Although \name\ slightly overshoots one of the trajectories, both forecast trajectories are plausible in this scenario. 
The heatmap in Fig.~\ref{fig:pred} (j) for these POIs shows a well coverage of the ground truth trajectories.

\section{Conclusion}
\label{sec:5}

In this paper, we have presented a novel method that incorporates all the three key elements in pedestrian trajectory prediction, namely social influence, multimodality, and scene constraints.
Our proposed method, \name, uses dynamic star graphs to model the social relationship between the POI and all other pedestrians in the scene during the observation period. 
In addition, the scene features are encoded through a semantic segmentation based scene encoder and the encoded scene features are merged with the social graph features to form the scene gated social graph features.
To generate multimodal predictions, trajectory samples are drawn through a latent variable whose parameters are learned from the scene gated social graph features.
The experimental results demonstrate that, with the two designed encoders and VAE module, \name\ achieves state-of-the-art performance on two widely used trajectory prediction datasets.
Finally, we observe that \name\ is simple yet effective as shown by the 
star graph structure and the memory usage experiments.

{\small
\bibliographystyle{ieee_fullname}
\bibliography{egbib}
}

\end{document}